# Latent Dirichlet Allocation Uncovers Spectral Characteristics of Drought Stressed Plants

**Mirwaes Wahabzada**[1*∘], **Kristian Kersting**[12*], **Christian Bauckhage**[1], **Christoph Römer**[2],
**Agim Ballvora**[3], **Francisco Pinto**[4], **Uwe Rascher**[4], **Jens Léon**[3], **Lutz Plümer**[2]
[1]Fraunhofer IAIS, Sankt Augustin, Germany. [2]Institute of Geodesy and Geoinformation, University of Bonn, Germany. [3]Institute of Crop Science and Resource Conservation, Plant Breeding, University of Bonn, Germany.
[4]Institute of Bio- and Geosciences, IBG-2: Plant Sciences, Forschungszentrum Jülich, Germany.

## Abstract

Understanding the adaptation process of plants to drought stress is essential in improving management practices, breeding strategies as well as engineering viable crops for a sustainable agriculture in the coming decades. Hyper-spectral imaging provides a particularly promising approach to gain such understanding since it allows to discover non-destructively spectral characteristics of plants governed primarily by scattering and absorption characteristics of the leaf internal structure and biochemical constituents. Several drought stress indices have been derived using hyper-spectral imaging. However, they are typically based on few hyper-spectral images only, rely on interpretations of experts, and consider few wavelengths only. In this study, we present the first data-driven approach to discovering spectral drought stress indices, treating it as an unsupervised labeling problem at massive scale. To make use of short range dependencies of spectral wavelengths, we develop an online variational Bayes algorithm for latent Dirichlet allocation with convolved Dirichlet regularizer. This approach scales to massive datasets and, hence, provides a more objective complement to plant physiological practices. The spectral topics found conform to plant physiological knowledge and can be computed in a fraction of the time compared to existing LDA approaches.

## 1 Introduction

Water scarcity is a principal global problem that causes aridity and serious crop losses in agriculture. It has been estimated that drought can cause a depreciation of crop yield up to 70% in conjunction with other abiotic stresses (Boyer, 1982; Pinnisi, 2008). Climate changes and a growing human population in parallel thus call for a sincere attention to advance research on understanding of plant adaptation under drought. A deep knowledge of the adaptation process is essential in improving management practices, breeding strategies as well as engineering viable crops for a sustainable agriculture in the coming decades. Accordingly, there is a dire need for crop cultivars with high yield and strong resistance against biotic and abiotic stresses.

*Both authors contributed equally. ∘Contact author: mirwaes.wahabzada@iais.fraunhofer.de

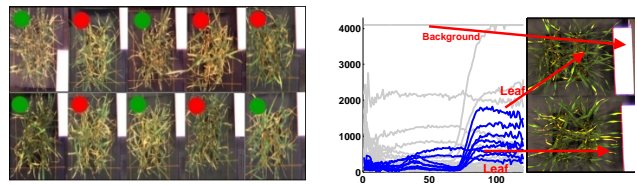

Figure 1: What are the specific spectral characteristics of plants suffering from drought stress? (Left) A collection of hyper-spectral images (projected to RGB space) within the flowering period. Stressed plants are indicated by red dots, control plants by green dots. Visually it is difficult to distinguish between control and stressed plant; compare e.g. the 2nd and 3rd image in the bottom row. (Right) Example spectral signatures taken from a hyper-spectral image for leaf and background pixels. (Best viewed in color)

Unfortunately, understanding stress is not an easy task. Stress resistance is the result of a complex web of interactions between the genotype and the environment leading to phenotypic expressions. It is contributed by a number of related traits that are controlled mostly by polygenic inheritance. In the past, a slow progress in the development of improving cultivars was mainly due to poor understanding of genetic factors that impact tolerance to drought (Passioura, 2002). Recently, progress has been made in under-standing the genetic basis of drought related quantitative trait loci (QTL), see e.g. (Lebreton et al., 1995; McKay et al., 2008). More recently, OMICS approaches have offered a direct molecular insight into drought tolerance mechanism, see e.g. (Rabbani et al., 2010; Guo et al., 2010; Abdeen et al., 2010). However, genetic and biochemical approaches are time consuming and still fail to fully predict the performance of new lines in the field. In recent years it is discussed that phenomic approaches, that measure the structural and functional status of plants may overcome the limited predictability and some authors have attributed this lack of high throughput phenomic data as the "phenomic bottleneck" (Richards et al., 2010).

Hyper-spectral imaging provides a particularly promising approach to sensor-based phenotyping. Its measurements were observed to contain early indicators of plant stress, see e.g. (Rascher et al., 2007; Rascher and Pieruschka, 2008). In contrast to conventional cameras, which record only 3 wavelengths per pixel, hyper-spectral cameras record a spectrum of several hundred wavelengths ranging from approximately 300nm to 2500nm resulting in big data cubes. These spectra contain information as to changes of the pigment composition of leaves which are the result of metabolic processes involved in plant responses to biotic or abiotic stresses. This information can be used e.g. using SVMs for classification of hyper-spectral signatures and in turn for prediction of biotic stress before symptoms become visible to the human eye, see e.g. (Rumpf et al., 2010; Römer et al., 2010), or for finding archetypical signatures (Kersting et al., 2012).

More important for the present study, hyper-spectral imaging was proven to be successful in discovering relationship between the spectral reflectance properties of vegetation and the structural characteristics of vegetation and pigment concentration in leaves (Govender et al., 2009); the spectral characteristics of vegetation are governed primarily by scattering and absorption characteristics of the leaf internal structure and biochemical constituents, such as pigments (e.g. chlorophyll a and b), water, nitrogen, cellulose and lignin, see (Curran et al., 1990; Gitelson and Merzlyak, 1996; Blackburn, 2007; Govender et al., 2009) and references in there). For instance, Gitelson and Merzlyak (1996) observed a increase of reflectance for the band 670nm if the amount of chlorophyll in the leaf was dropped as it is case in the presence of drought stress.

However, since stress reactions are the result of a complex web of interactions between the genotype and the environment, indices involving very few distinct wavelengths only run the risk of providing a oversimplified and actually wrong spectral characterization of drought reactions. In this study, we present a data-driven approach to discovering spectral drought stress indices, treating it as an unsupervised labeling problem at massive scale and solving it using an online variational Bayes approach (Hoffman et al., 2010; Wahabzada and Kersting, 2011) to latent Dirichlet allocation (Blei et al., 2003). Although, there are other data-driven approaches, such as low-rank matrix factorization based on sub-sampling (Mahoney and Drineas, 2009; Sun et al., 2008) or computing extreme data points (Thurau et al., 2012), these methods have a major limitation: they are not part-based. Part-based methods such as NMF, however, do not easily deal with additional information, e.g. relational information and short range dependencies.

Using a topic model is a sensible idea since it is common practice in plant physiology to talk about integral wavelength-reflectance pairs only (the words) within signatures (the documents). However, we have to be a little bit more careful. Signatures are still "curves" showing important short range dependencies among spectral wavelengths. In order to preserve the dependencies, we develop an online variational Bayes approach with convolved Dirichlet regularizer (Newman et al., 2011). Indeed, one may argue that subsampling is a valid alternative to deal with the massive amount of data at hand. However, plant physiologists often do not get used to the idea of throwing away information and actually do not trust the results of sample-based methods. In contrast, our regularized online VB scales well to massive datasets and does not throw away information, hence, provides a more objective complement to plant physiological practices. Moreover, the spectral topics found by LDA conform to plant physiological knowledge and can be computed in a fraction of the time compared to existing LDA approaches.

We proceed as follows. We start off by reviewing online variational Bayes for LDA. Afterwards, we develop the regularized variant. Before concluding we will present our main experimental evaluation on hyper-spectral images of plants with two treatments (control and stressed) as well as supplemental evaluations on two additional real world datasets, the network of human diseases and Wikipedia articles.

## 2 Online Variational Bayes for LDA

LDA is a Bayesian probabilistic model of collections of text documents (Blei et al., 2003). It assumes a fixed number of $K$ underlying topics in a document collection. Topics are assumed to be drawn from a Dirichlet distribution, $\beta_k \sim Dir(\eta)$, which is a convenient conjugate to the multinomial distribution of words appearing in documents. According to LDA, documents are generated by first drawing topic proportions according to $\theta_d \sim Dir(\alpha)$, where $\alpha$ is the parameter of the Dirichlet prior on the per-document topic distributions. Then for each word $i$ a topic is chosen according to $z_{di} \sim Mult(\theta_d)$ and the observed word $w_{di}$ is drawn from the selected topic, $w_{di} \sim Mult(\beta_{z_{di}})$.

In this paper, we focus on variational Bayesian (VB) inference. Here, the true posterior is approximated using a simpler, fully factorized distribution $q$. Following Blei et al. (2003); Hoffman et al. (2010), we choose $q(z, \theta, \beta)$ of the form $q(z_{di} = k) = \phi_{dw_{di}k}$, $q(\theta_d) = Dir(\theta_d, \gamma_d)$, and $q(\beta_k) = Dir(\beta_k, \lambda_k)$. The variational parameters $\phi$, $\gamma$, and $\lambda$ are optimized to maximize the Evidence Lower BOund (ELBO)

$$\log p(w \mid \alpha, \eta) \geq \mathcal{L}(w, \phi, \gamma, \lambda)$$
$$\triangleq \mathbb{E}_q \left[ \log p(w, z, \theta, \beta \mid \alpha, \eta) \right] - \mathbb{E}_q \left[ \log q(z, \theta, \beta) \right],$$

which is equivalent to minimizing the Kullback - Leibler divergence between $q(z, \theta, \beta)$ and the true posterior $p(z, \theta, \beta \mid w, \alpha, \eta)$.

Based on VB, Hoffman et al. (2010) have introduced an online variant that we here present for the batch case running over mini-batches (chunks of multiple observations). That is, we assume that the corpus of documents has been sorted uniformly at random and chunked into $l$ mini-batches $B_1, B_2, \ldots, B_l$ of size $S$. That is, the ELBO $\mathcal{L}$ is set to maximize $\mathcal{L}(w, \phi, \gamma, \lambda) \triangleq \sum_{B_i} \sum_{d \in B_i} \ell(n_d, \phi_d(n_d, \lambda), \gamma_d(n_d, \lambda), \lambda)$, where $n_d$ is the word count vector and $\ell(n_d, \phi_d(n_d, \lambda), \gamma_d(n_d, \lambda), \lambda)$ denotes the contribution of document $d$ to the ELBO. As Hoffman et al. (2010) have shown this mini-batch VB-LDA corresponds to a stochastic natural gradient algorithm on the variational objective $\mathcal{L}$. Using mini-batches reduces the noise in the stochastic gradient estimation as we consider multiple observations per update: $\tilde{\lambda}_{kw} = \eta + D/S \sum_{s \in B_i} n_{sw} \phi_{skw}$ where $n_{sw}$ is the $s$-th document in the $i$-th mini-batch and $D$ denote the number of documents. The rate of change $\rho_t$ is set to $\rho_t \triangleq (\tau_0 + t)^{-\kappa}$ where $\tau_0 \geq 0$ and $\kappa \in (0.5, 1]$.

## 3 Regularized Variational Bayes

For introducing a structured prior to regularize the word-topic probabilities, we are inspired by the recent regularized Gibbs (regGS) approach due to Newman et al. (2011), who have demonstrated that regularization improves the topic coherence. Specifically, we view each topic as a mixture of word probabilities given by the word-pair dependency matrix $C$ (a $W \times W$ matrix, where $W$ denotes the size of vocabulary and $C_{ij} \geq 0$), that is

$$\beta_k \propto Cb_k \text{ where } b_k \sim Dir(\eta), \quad (1)$$

In VB the true posterior is approximated using fully factorized distributions $q$. Consequently, we parametrize the word probabilities $b$ by introducing a new variational parameter $\nu$, i.e. $q(b_k) = Dir(b_k, \nu_k)$. The per-word topic assignments $z$ are parametrized by $\phi$, and the posterior over the per-document topic weights $\theta$ are parametrized by $\gamma$, as for the standard LDA (Blei et al., 2003). With this, the part of the likelihood including the specific parameter $\nu$ can be written as

$$\mathcal{L}_{[\nu]} = \mathbb{E}_q[\log p(w \mid z, C, b)]$$
$$+ \mathbb{E}_q[\log p(b \mid \eta)] - \mathbb{E}_q[\log q(b)]. \quad (2)$$

The remaining part of the ELBO does not change. To approximate the first term of Eq. (2) we adapt the lower bound on the log-sum-exp function (Boyd and Vandenberghe, 2004, page 72), $\mathbb{E}_q[\log \sum_i X_i] \geq \log \sum_i \exp(\mathbb{E}_q[\log X_i])$ (for a detailed proof see e.g. (Paisley, 2010)) to our case, which follows by applying Jensen's inequality: $\mathbb{E}_q[\log p(w \mid z = k, C, b)]$

$$= \sum_i^W \Phi_{ik} \mathbb{E}_q[\log \sum_j^W C_{ij} b_{jk}]$$
$$\geq \sum_i^W \Phi_{ik} \log \sum_j^W \exp(\mathbb{E}_q[\log C_{ij} b_{jk}])$$
$$= \sum_i^W \Phi_{ik} \log \sum_j^W C_{ij} \exp(\mathbb{E}_q[\log b_{jk}]) \quad (3)$$

where $\sum_w^W \Phi_{wk} = \sum_d^D \sum_w^W \phi_{dwk}$. This is still a lower bound, so maximizing it will improve the ELBO. The expectation of $\log b$ under the distribution $q$ is: $\mathbb{E}_q[\log b_{wk}] = \Psi(\nu_{wk}) - \Psi(\sum_s \nu_{sk})$. The remaining terms of the Eq. (2) (for a topic k) are

$$\mathbb{E}_q[\log p(b \mid \eta)]_{[k]} = \log \Gamma(W \eta) - W \log \Gamma(\eta)$$
$$+ \sum_w^W (\eta - 1)(\Psi(\nu_{wk}) - \Psi(\sum_s^W \nu_{sk})),$$
$$\mathbb{E}_q[\log q(b)]_{[k]} = \log \Gamma(\sum_s^W \nu_{wk}) - \sum_w^W \log \Gamma(\nu_{wk})$$
$$+ \sum_w^W (\nu_{wk} - 1)(\Psi(\nu_{wk}) - \Psi(\sum_s^W \nu_{sk})).$$

To derive a VB approach, we compute the derivative of Eq. (2) with respect to the variational parameter $\nu_{wk}$. After applying the chain rule and rearranging terms, this gives e.g. $\partial \mathbb{E}_q[\log p(w \mid z, C, b)]/\partial \nu_{wk} =$

$$= \Psi_1(\nu_{wk}) \sum_i \Phi_{ik} \frac{C_{iw} \exp(\mathbb{E}_q[\log b_{ik}])}{\sum_j C_{ij} \exp(\mathbb{E}_q[\log b_{jk}])}$$
$$- \Psi_1(\sum_s v_{sk}) \sum_i \Phi_{ik}.$$

Taking the derivatives for all terms together we arrive at: $\partial \mathcal{L}/\partial \nu_{wk} =$

$$\Psi_1(\nu_{wk})(\sum_i^W \Phi_{ik} \frac{C_{iw} \exp(\mathbb{E}_q[\log b_{ik}])}{\sum_j^W C_{ij} \exp(\mathbb{E}_q[\log b_{jk}])} + \eta - \nu_{wk})$$
$$- \Psi_1(\sum_s^W \nu_{sk}) \sum_i^W (\Phi_{ik} + \eta - \nu_{ik}). \quad (4)$$

Setting the above derivative to zero, we get the following fixed point update:

$$\nu_{wk} = \eta + \sum_{i}^{W} \Phi_{ik} \frac{C_{iw} \exp(\mathbb{E}_q[\log b_{ik}])}{\sum_{j}^{W} C_{ij} \exp(\mathbb{E}_q[\log b_{jk}])} . \quad (5)$$

This is a proper generalization of the standard VB approach. To see this simply set the word-pair dependency matrix $C$ to the identity matrix. Then, it follows that $\nu_{kw} = \eta + \sum_d n_{dw}\phi_{dwk}$; see also (Blei et al., 2003) for more details.

To derive a learning algorithm, i.e., to actually optimize $\mathcal{L}$, we follow a coordinate ascent on the variational parameters $\phi$, $\gamma$ and $\nu$. Given the word topic probabilities $\beta$ from Eq. (1), this yields the following per-document updates for $\phi$ and $\gamma$ in the E-step:

$$\phi_{dwk} \propto \beta_{wk} * \exp(\mathbb{E}_q[\log \theta_{dk}]) , \quad (6)$$

$$\gamma_{dk} = \alpha + \sum_{w}^{W} n_{dw}\phi_{dwk} . \quad (7)$$

In the M step, we perform fixed point updates, as given in Eq. (5), and compute the values $\beta_{wk}$ using Eq. (1) as follows:

$$\beta_{wk} \propto \sum_{i}^{W} C_{iw} \exp(\Psi(\nu_{ik}) - \Psi(\sum_{s}^{W} \nu_{sk})) . \quad (8)$$

However, recall that one of our main goals is the application of regularised VB to hyper-spectral images of plants. Since a single image can already consists of hundreds of thousands of signatures (documents) so that several images (as in our experiments) easily scale to several million documents, batch VB is likely to be overtaxed in terms of running time. Consequently, we will develop an online variant of regularized VB (regVB) that scales well to massive datasets.

## 4 Online Regularized VB

Since setting the word-dependency matrix $C$ to identity matrix results in standard VB, it is intuitively clear that we can extend the regularized VB to the online case (regOVB) by adapting (Hoffman et al., 2010). Specifically, the variational lower bound for the regVB can be written as $\mathcal{L} =$

$$\sum_{d}^{D} \left\{ \mathbb{E}_q[\log p(w_d \mid \theta_d, z_d, C, b)] + \mathbb{E}_q[\log p(z_d \mid \theta_d)] \right.$$
$$-\mathbb{E}_q[\log q(z_d)] + \mathbb{E}_q[\log p(\theta_d \mid \alpha)] - \mathbb{E}_q[\log q(\theta_d)]$$
$$\left. + (\mathbb{E}_q[\log p(b \mid \eta)] - \mathbb{E}_q[\log q(b)])/\mathbf{D} \right\}$$
$$\triangleq \sum_{d}^{D} \ell(n_d, \phi_d, \gamma_d, \boldsymbol{C}, \boldsymbol{\nu}) , \quad (9)$$

where $\ell(n_d, \phi_d, \gamma_d, \boldsymbol{C}, \boldsymbol{\nu})$ is the $d$th document's contribution to the variational bound. The per-corpus terms

---

**Algorithm 1**: Online regularized LDA. The changes to online LDA are highlighted using blue fonts.

**Input**: $D$ (documents), $S$ (batchsize), $RegIter$ (fixed updates in M step), $C$ (word-dependency matrix)

Define $\rho_t \triangleq (\tau_0 + t)^{-\kappa}$ with $\kappa \in (0.5, 1]$;
Initialize $\boldsymbol{\nu}$ randomly and set $t = 0$;
**repeat**
    Select $S$ documents randomly forming the mini-batch $\tilde{D}$;
    /* Compute E step */
    **foreach** *document d in $\tilde{D}$* **do**
        **repeat**
            Set $\phi_{dwk} \propto \beta_{wk} * \exp(\mathbb{E}_q[\log \theta_{dk}])$;
            Set $\gamma_{dk} = \alpha + \sum_{w}^{W} \phi_{dwk} n_{dw}$;
        **until** $\frac{1}{K}\sum_k |change\ in\ \gamma_{dk}| < 0.00001$ ;
    /* Compute M step */
    Initialize $\tilde{\nu}$ randomly;
    **for** $i = 1 : RegIter$ **do**
        $\tilde{\nu}_{wk} = \frac{D}{S}\sum_{d\in\tilde{D}}\sum_{i}^{W} \phi_{dik} \frac{C_{iw}\exp(\mathbb{E}_q[\log b_{wk}])}{\sum_j C_{ij}\exp(\mathbb{E}_q[\log b_{jk}])} + \eta$;
    Set $\boldsymbol{\nu} = (1-\rho_t)\boldsymbol{\nu} + \rho_t\tilde{\boldsymbol{\nu}}$;
    $\beta_{wk} \propto \sum_{i}^{W} C_{iw}\exp(\Psi(\nu_{ik}) - \Psi(\sum_s \nu_{sk}))$;
    Increment $t := t + 1$;
**until** *converged* ;

---

are summed together and divided by the number of documents $D$. Doing so allows one to derive the online approach since the optimal $\nu$ is the one for which $\mathcal{L}$ maximized after fitting the per-document parameter. In other words, we can use the regularized updates in a per-document manner. This is summarized in Alg. 1. That is, we start off by randomly selecting documents form entire dataset by forming a mini-batch $\tilde{D}$. Then an *E step* is performed to find locally optimal values of $\gamma$ and $\phi$ holding $\beta$ fix. In the *M step* several fixed point updates for $\tilde{\nu}$ are computed using $\tilde{\nu}_{wk} =$

$$\frac{D}{S} \sum_{d\in\tilde{D}} \sum_{i} \phi_{dik} \frac{C_{iw}\exp(\mathbb{E}_q[\log b_{ik}])}{\sum_j C_{ij}\exp(\mathbb{E}_q[\log b_{jk}])} + \eta \quad (10)$$

given the document-specific parameter $\phi_d$ with $d \in \tilde{D}$ (currently observed mini-batch), where we rescale by $\frac{D}{S}$ to update as though we would have seen all documents. Multiple documents are used per update to reduce variance. The parameter $\boldsymbol{\nu}$ is updated through a weighted average of its previous value, and $\tilde{\nu}$ (computed for the current mini-batch using fixed point updates as in Eq. (10)). Furthermore, the new values of $\beta$ are computed given $\boldsymbol{\nu}$ and word-dependency matrix $C$. Following Hoffman et al. (2010), the rate of change $\rho_t$ is set to $\rho_t \triangleq (\tau_0 + t)^{-\kappa}$ with $\kappa \in (0.5, 1]$ in order to guarantee convergence. Note, as in the non-regularized case, we recover regularized batch VB when setting the batch size to $S = D$ and $\kappa = 0$.

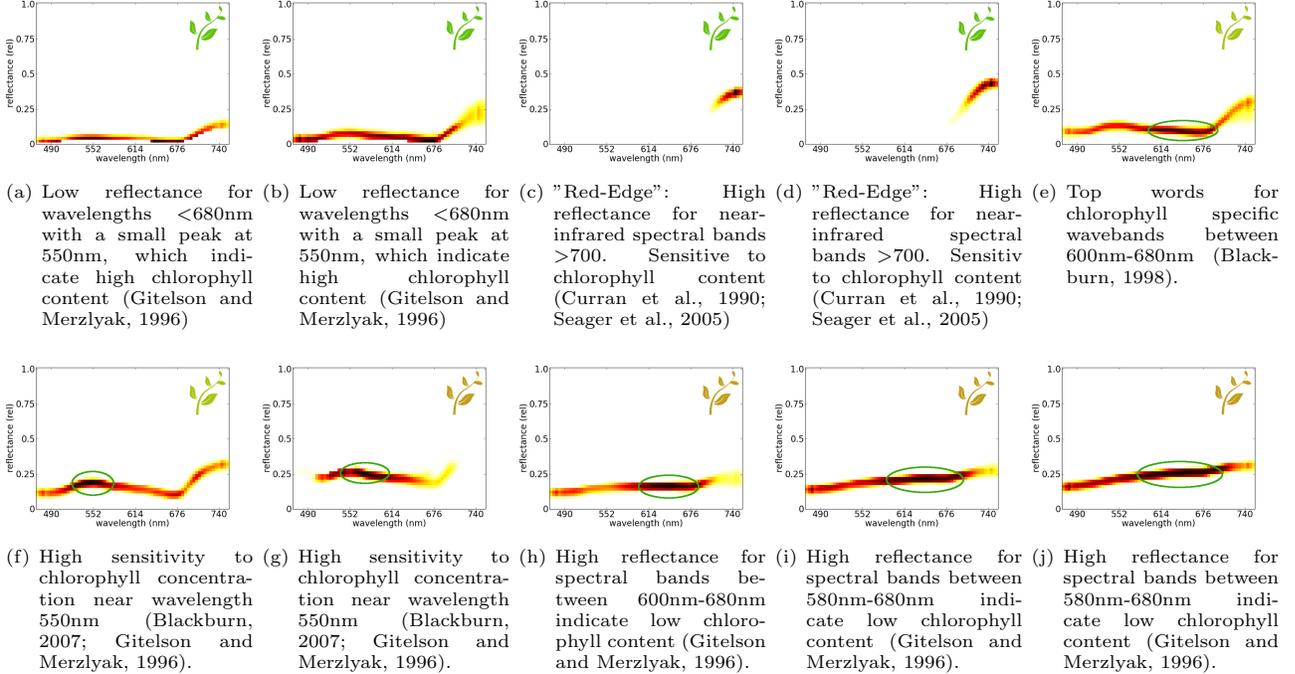

Figure 2: Example topics learned by regularized online VB (K = 15). Most of the topics can be found to reflect results received by experts (see Table 1 in (Govender et al., 2009) and references in there). Top words (here in dark red) consists mostly of highly correlated spectral bands. Furthermore, given the reflectances one can identify whether a topic represent a healthy (e.g. low reflectance in chlorophyll specific bands, large reflectance in near infrared wavebands (Blackburn, 2007)) or non-healthy signature. (Best viewed in color)

## 5 Uncovering Spectral Drought Stress Characteristics

In our main experiment, drought stress was applied to barley cultivar Scarlett. Hyper-spectral images of the 5 stressed and 5 control Barleys were taken with a resolution of 640x640 pixels, where each pixel is a vector with 120 recorded wavelengths from the range of 394-891nm with approximately 4nm spectral resolution, using the SOC-700 hyper-spectral imaging system (Rascher et al., 2007), manufactured by Surface Optics. A normalization of the images was done by calculating the spectral reflectance for each pixel. For that, the spectrum of a pixel was divided by the spectrum of the incoming radiation estimated from a white reference panel that exhibits Lambertian reflectance located in each scene. Because the monitoring started with the flowering time of Barley, both the control and stressed Barleys developed senescent leaves. In our experiments we used 7 measurements for 10 plants, which were done every 3-4 day. Five control plants grew in a fully water capacity of the substrate conditions while the 5 stressed plants were exposed to 50% water supply reduction (at BBCH 30; BBCH is a scale used to identify the phenological development stages of a plant). This yielded 70 data cubes of resolution 640x640x120. Although the SOC-700 measured from 394nm to 890nm, the wavelengths below 470nm and above 750nm were discarded because they appeared to be very noisy. The reason for this is most likely an unstable source of illumination for these frequencies. Therefore, only the bandwidths from 470nm to 750nm were used. We transform each cube into a dense 'pixel x spectra' matrix. This resulted in 70 data matrices of soze 69x409,600; overall a dataset with about 30 Million signatures.

Furthermore, for our analysis we are interested in finding not just specific wavelength patterns but also reflectance values. For that reason, and to get the data sparse, we discretize the corresponding signatures in the following way: we decompose the space covering the full signatures (with $Wl = 69$ spectral bands) additionally into $R = 50$ possible reflectance words. Thus, in a signature each wavelength can consist of one of the $R$ distinct reflectance words, which results in a total number of $Wl \times R$ different possible *spectral words*. The use of discretised values instead of continuous is also motivated by the fact that according to plant physiologist small difference in reflectance values are not of great importance. They instead study the spectral characteristics of plants in order to get spectral indices more sensitive to pigment content. Furthermore, we excluded all biologically implausible signatures ("background" signatures) using Simplex Volume Maximization (SiVM) (Thurau et al., 2012) algorithm, by taken out all signatures with a high coefficient to an extreme spectra which was identified as "background" (non-leaf). This resulted in a dataset with about $9,2$ Million signatures (documents).

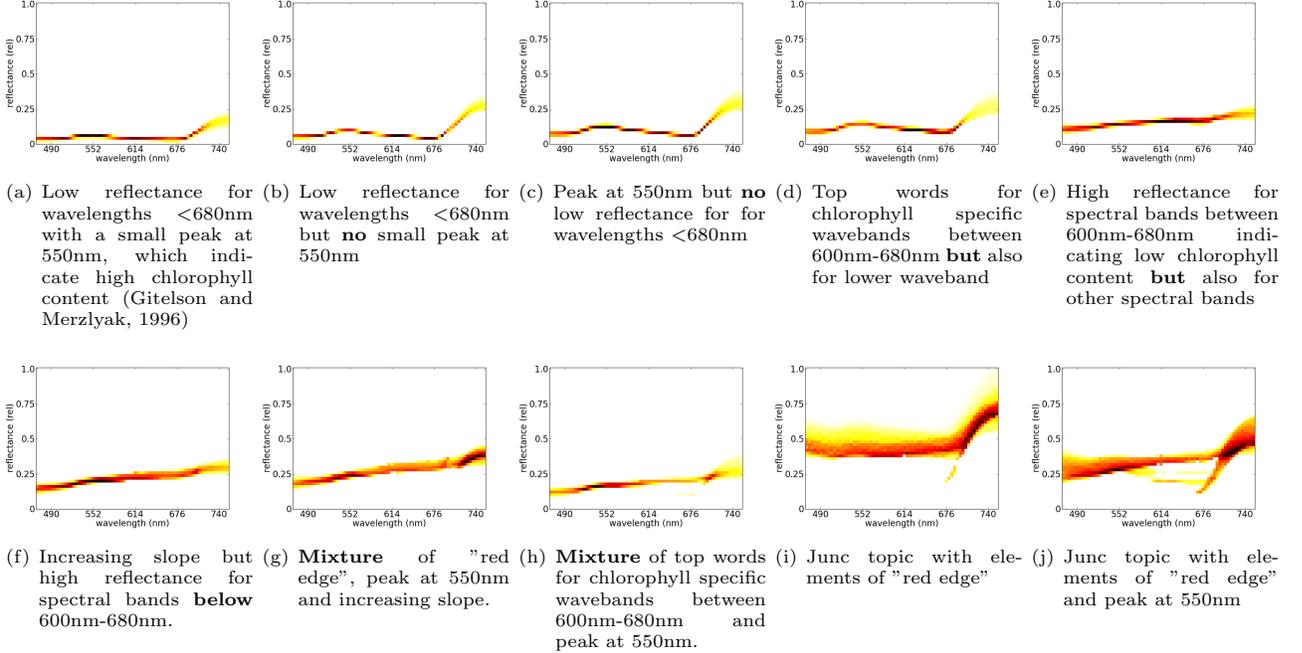

Figure 3: Topics learned by standard online VB (K = 15). As one can see the topics are less distinct compared to the regularized ones and actually mix known indices in particular for stressed and controlled plants.

The word-dependency matrix C was created using pointwise mutual information (PMI) (Newman et al., 2011). For plants, in order to also get the cooccurrences between different reflectance's within a wavelength, we proceeded as follows: we first aggregated the signatures in the images within each non-overlapping squares of 5x5 pixel. *Spectral word* cooccurrences were computed using a sliding window of length 1 in each direction (wavelength and reflectances) in the aggregated signatures. We compute the PMI only for the 1000 words with the highest frequency. The matrix C was computed from 8 plants (56 images, with approx. $7,3$ Million signatures), for the remaining two plants (one control and one stressed, with approx. $1,9$ Million signatures) an regularized/non-regularized online LDA was learned on non aggregated signatures. For both methods we set the batchsize $S = 1024$, $\kappa$ close to 0.5, $\alpha = 0.01$, $\eta = 0.01$ and (for the regularized online VB) 10 fixed point iterations in the M step. For both methods we stopped when each signature (document) was seen once. The computation took for both methods less than one hour (for $K = 15$) on a standard Intel-Quadcore 3.4 GHz computer, but only using a single core. In contrast, running batch LDA on two plants took more than 15 hours.

### 5.1 Uncovered Characteristics

Fig. 2 shows examples of topics discovered by regularized online VB (K = 15). The topics conform to common plant physiological knowledge (see also Table 1 in (Govender et al., 2009) and references in there). For example topics (a), (b), (f) and (g) clearly show that high probable words appear for the spectral band 550nm. This wavelength was found to have a maximum sensitivity to a wide range of chlorophyll contents (Blackburn, 2007; Gitelson and Merzlyak, 1996), which are the most important pigments as they are necessary for photosynthesis. Furthermore, topics (a) and (b) show a high probability wavelengths close to 500nm and 670nm. As mentioned by Gitelson and Merzlyak (1996), these bands/wavelength have high correlation for yellow-green to dark-green leaves. Additionally Gitelson and Merzlyak report that the reflectance in 550nm does not exceed 0.1 in green leaves (for Maple and Chestnut). This is also supported by topics (a) and (b) but now for Barley. An up-rising slope between bands 690nm and 750nm is known as the "red edge" and is due to the contrast between the strong absorption of chlorophyll and the otherwise reflective leaf (Seager et al., 2005). The "red edge" is detected by topics (c) and (d). The discovered important wavelengths in the topics (h)-(j) (between 580nm-680nm) also closely mirror known indices. Blackburn (1998) define the "optimal" individual bands for pigment estimation as 680 nm for chlorophyll a and 635 nm for chlorophyll b. Gitelson and Merzlyak (1996) observed a increase of reflectance for the band 670nm if the amount of chlorophyll in the leaf drops, as it is the case for stressed plants. Taking both results together, topics (h)-(j) clearly describe "non-healthy" resp. "drought-stressed" topics and also conform to plant physiological knowledge.

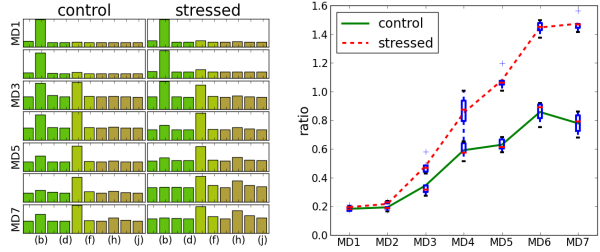

Figure 4: (**Left**) Evolution of "non-healthy" and "healthy" topics shown in Fig.2 over time. Each row stands for different measurement days 1-7. The larger the bar the more prominent a topic is. As one can see, the most prominent topics for control plants consists of "healthy" yellow-green topics, whereas for the stressed plants the probabilities of "healthy" topics drop rapidly and "stress" topics (brown) become more likely in latter days. (**Right**) The ratio of relative importance of a "non-healthy" and "healthy" topic. Beginning by measurement day 3 we have a significant difference (paired t-test, p = 0.05) between stressed and control plants. (Best viewed in color)

The discovered topics/indices can be used in various ways to investigate drought stress reactions. For instance, we can computed the distributions of topics for plants over time. Fig. 4 (**left**) shows this topic distribution over all the measurement days 1-7 (recall that images were taken every 2-3 days so we cover several weeks). Since, LDA is giving us a topic distribution per signature only, we actually estimated the concentration parameters of the Dirichlet distribution over the topics induced by all signatures of single images (Minka, 2000). Then, we used the expected probability of each topic for visualization.

As one can see, the probability of "non-healthy" topics increases for the stressed plant over time, whereas for the control plant "healthy" yellow-green to green topics have higher weights. Furthermore, Fig. 4 (**right**)

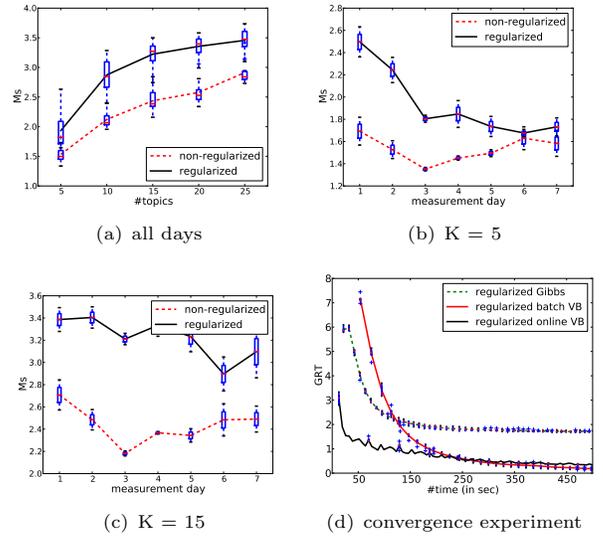

Figure 5: Regularization produces more specific topics. Shown are the averages per topic KL-distance to a "background topic" when taking (a) all days as a function of the number of topics, (b) for 5 topics and (c) 15 topics as a function of measurement days. The topic proportions of the signatures from each images were taken separately in order to compute the distance to "background topic". As one can see the regularized topics are more specific than non-regularized ones; they have significantly larger KL-distance. Interestingly, for the last three measurement days the picture gets more diverse. Actually, regularized topics capture the spectral characteristics less well for smaller (b) than for large number of topics (c). This indicates that their is a diverse set of spectral characteristics required to capture drought stress. Furthermore, (d) regularized VB converges significantly faster than regularized semi-collapsed Gibbs-Sampling (Wikipedia, K = 20). Shown are the expected change of word probabilities in topics over time. This speed-up puts high throughput hyper-spectral topic models of several plants per day for several weeks in reach. (Best viewed in color)

show the ratio between the expected probability of a "non-healthy" and "healthy" topic (ratio = h/b, for topics (b) and (h) in Fig. 4 (**left**)). For this experiment we also computed the distribution of topics for all plants and measurement days using the model shown in Fig. 2. Beginning by measurement day 3 this ratio shows a significant difference (paired t-test, p = 0.05) between stressed and control plants. This are yet another validations that the discovered topics capture drought stress relevant characteristics.

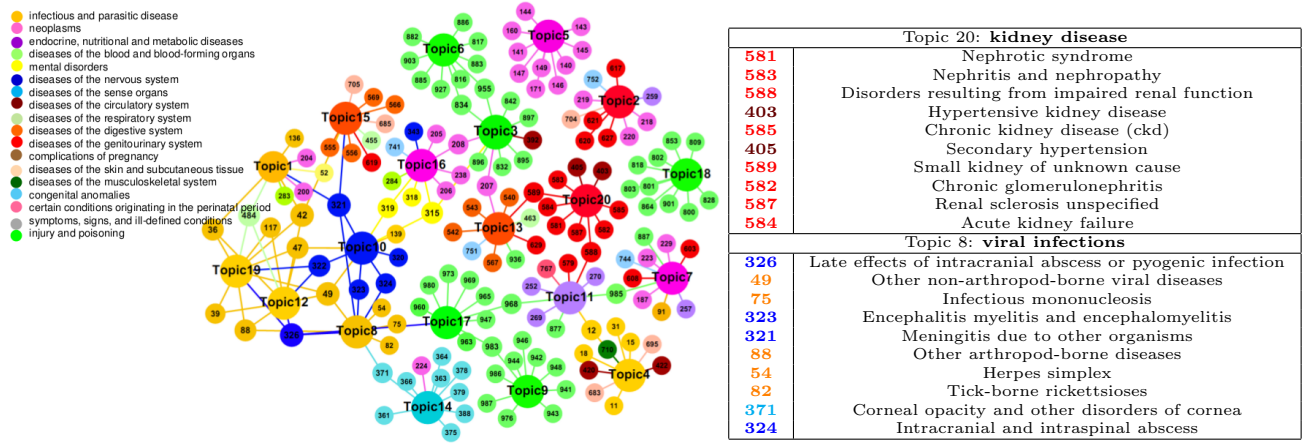

Figure 6: Regularized VB LDA can discover composite topics in Human Disease Network: (Left) A graph representing topics learned by regularized VB (K = 20). Topics (big circles) are connected with top 10 diseases (small circles). The different colors indicate different categories of ICD9 codes. As one can see, most of the topics are dominated by one of the categories. But there are also mixed topics of common diagnosed diseases across several categories. (Right) Two example topics learned with regularized VB LDA about "kidney diseases" and "viral infections". (Best viewed in color)

### 5.2 Regularization improves the topic coherence

To accommodate the qualitative results so far with quantitative ones, we compared the results produced by regularized and non-regularized oLDA in terms of their topic coherence. A topic is specific or has authentic identity if it is far from generating words in wide range of documents. To measure this, Alsumait et al. (2009) define a "background topic" to be a topic that is equally probable in all the documents, i.e., $P(d_m \mid \xi) = 1/D$ for $m \in (1, 2, ..., D)$. In turn, the distance between a topic and the "background topic" can be viewed as measure for how much background or common the information is provided by the topic. Since we are interested in overall performance of a model, we measured the average per topic distance to the "background topic" using KL-Divergence:

$$Ms = \frac{1}{K} \sum_k D_{kl}(\boldsymbol{\theta}_k, \xi) \text{ where } \boldsymbol{\theta}_k = (\theta_{1k}, \theta_{2k}, ..., \theta_{Dk}) .$$

The results are summarized in Fig.5 (a)-(c). As one can see, the regularized topics are more specific than non-regularized ones (a), i.e. they have larger KL-distance . More interestingly, for the last three measurement days regularized online LDA does worst for lower number of topics (b) than for large number of topics (c). This can be due to wide range of different signatures in the images (in terms of "non-healthy" signatures). This can be also seen in Fig.4 where for the first days there just a few topics with high probability, whereas for the latter days more topics are needed to represent the data.

## 6 Supplementary Evaluation

To further investigate the performance of regularized VB works even if LDA fails to learn anything meaningful, and that it is comparable with semi-collapsed Gibbs Sampling inference by Newman et al. (2011) we provide experiments with two additional datasets.

**Human Disease Network:** Here we investigate the question: can regularized VB learn meaningful and interpretable topics models even if "no information" in documents are available? To answer this we make use of human disease network dataset (Hidalgo et al., 2009), which consist originally of approximately 32 Million inpatient claims, pertaining 13039018 individuals of 65 years and older patients. The corresponding human disease network includes cooccurrences of diagnosis (specified by ICD9 codes) and up to 9 secondary diagnosis.

Here we applied regularized VB by using disease dependency matrix which was computed using PMI. We used only diseases with more than 500 occurrences, resulted in a total of 742 diseases. In order to learn a regularized topic model, we created a synthetic datasets of D = W documents each consisting of one word (actually an identity matrix with $W \times W$ dimensionality). We used the following settings: $\alpha = 0.01, \eta = 0.01$ and 10 fixed point iterations in the M step. The results of the regularized VB (for K = 20) are shown in Fig.6. As one can see, regularized VB LDA can discover coherent topics in human disease network, even if LDA would fail. Topics (big circles) are connected with top 10 diseases (small circles). The different colors indi-cate different categories of ICD9 codes. As one can see, most of the topics are dominated by one of the categories. But there are also mixed topics of common diagnosed diseases across several ICD9 categories, as shown in the table on Fig.6, with topic about "kidney diseases" and "viral infections".

**Wikipedia Dataset:** Further, to compare regularized VB with regularized Gibbs Sampling we used a small set of D = 5000 Wikipedia articles, with W = 7312 words in the vocabulary and N ≈ 750000 total number of terms. The word cooccurrences were computed using an extern data of 3441010 titles of Wikipedia articles. The titles include naturally the short range word dependencies of words. Here we used the normalized cooccurrences (matrix $C$) for all regularized methods. We set $\alpha = 0.05 \frac{N}{DK}$ (as suggested by Newman et al. (2011)), $\eta = 0.01$ and the number of topics was set to K = 20, and in each M step 10 fixed point iterations were applied. For the online case the batchsize was set to S = 500, $\kappa$ close to 0.5 and $\tau_0 = 1024$. The regularized Gibbs LDA was run for 1250 iterations where we applied regularization (10 fixed point updates) every 50 iterations. The VB methods were run until each document was seen 50 times. The results are shown in Table 1. As one can see, all methods produce topic models of similar quality in terms of the interpretability. Additionally, to show convergence of the different methods, we computed the change of the probabilities of words $\boldsymbol{\beta}$ between two iterations when $\boldsymbol{\beta}$ was computed. To measure this, we use: $GRT = \sum_w \sum_k \mid \beta_{kw}^t - \beta_{kw}^{t-1} \mid$, where t indicate the current iteration. The results are represented in Fig.5 (d). As one can see, the regularized batch VB converges faster than semi-collapsed Gibbs-Sampling. Moreover, regularized online VB outperforms regularized batch VB and is comparable in terms of the interpretability of the topics.

## 7 Conclusion

Understanding drought stress in plants is not an easy task. In this context, hyper-spectral image sensors are an established, sophisticated method for discovering spectral stress indices. However, they gather massive, high dimensional data clouds over time, which together with the demand of physical meaning of the prediction model present unique computational problems in scale and interpretability. Motivated by this, we developed a regularized variational Bayes approach to latent Dirichlet allocation and presented the — to the best of our knowledge — the first application of probabilistic topic models to discovering drought stress characteristics from hyper-spectral image sequences. Our experimental results on a large-scale plant phenotyping dataset demonstrate that the estimated spectral char-acteristics are meaningful, conform to existing plant physiological knowledge, and are fast to compute. In contrast to indices established in plant physiology, topics are not based on single or few wavelengths but provide a distributional view on the characteristics of complete signatures. Overall, our results are an encouraging sign that the vision of high throughput precision phenotyping is not insurmountable. Detailed measurements of plant characteristics can be analysed at massive scale to collectively provide estimates of trait phenotypes for many of the underlying genotypes that comprise a typical plant breeding population.

|  | music |  |
|---|---|---|
| **regGS** | music, band, song, released, live, new, single, rock, songs, records | + |
| **regVB** | music, band, song, released, live, new, songs, single, rock, records | + |
| **regOVB** | music, band, song, released, single, songs, rock, live, records, track | + |
|  | **league** |  |
| **regGS** | league, team, season, game, first, years, games, club, two, new | + |
| **regVB** | league, team, club, world, cup, years, first, national, won, season | + |
| **regOVB** | league, season, team, club, years, new, state, career, played, born | + |
|  | **university** |  |
| **regGS** | university, new, american, college, school, science, research, professor, national, born | + |
| **regVB** | university, new, united, born, states, american, national, school, first, party | + |
| **regOVB** | university, book, research, professor, science, published, work, new, first, school | + |
|  | **church** |  |
| **regGS** | school, high, church, new, schools, district, year, students, education, college | + |
| **regVB** | church, war, king, first, century, great, new, history, city, catholic | + |
| **regOVB** | church, house, building, century, city, town, new, village, built, old | + |

Table 1: Gibbs, batch VB and online VB for regularized LDA produce qualitatively comparable topics: An example of topics (for Wikipedia Dataset) represented by 10 words with highest weights (K = 20).

Our work provides several interesting avenues for future work. Next to experiments under field conditions e.g. in an experimental agricultural site, one should aim at improving the topics quality even further by applying hierarchical, (semi-)supervised and relational versions of topic models. Active LDA approaches could speed up computations even further. Ultimately, the models should be used to identify the most relevant moment when biologists have to gather samples for invasive, molecular examinations.

**Acknowledgements:** The authors thank the anonymous reviewers for their valuable comments, Edwin Bonilla, Wray Buntine, and Zhao Xu for helful discussions on reglurarized LDA, and Anja Pilz and Hannah Lanzrath for helping with the data for the supplementary evaluation. The work was partially supported by the Fraunhofer ATTRACT fellowship STREAM and by the German Federal Ministry of Education and Research BMBF/315309/CROP.SENSe.

# References

A. Abdeen, J. Schnell, and B. Miki. Transcriptome analysis reveals absence of unintended effects in drought-tolerant transgenic plants overexpressing the transcription factor abf3. *BMC Genomics*, 11, 2010.

L. Alsumait, D. Barbará, J. Gentle, and C. Domeniconi. Topic significance ranking of lda generative models. In *Proceedings of ECML*, pages 67–82, 2009.

G. A. Blackburn. Quantifying chlorophylls and carotenoids at leaf and canopy scales: an evaluation of some hyperspectral approaches. *Remote Sens. Environ*, 66:273–285, 1998.

G. A. Blackburn. Hyperspectral remote sensing of plant pigments. *Journal of Experimental Botany*, 58(4):855–867, 2007.

D.M. Blei, A. Ng, and M. Jordan. Latent dirichlet allocation. *Journal of Machine Learning Research*, 3:993–1022, 2003.

S. Boyd and L. Vandenberghe. *Convex Optimization*. Cambridge University Press, New York, NY, USA, 2004.

J.S. Boyer. Plant productivity and environment. *Science*, 218:443–448, 1982.

P.J. Curran, J.L. Dungan, and H.L Gholz. Exploring the relationship between reflectance red edge and chlorophyll content in slash pine. *Tree Physiology*, 7:33–48, 1990.

A.A. Gitelson and M.N. Merzlyak. Signature analysis of leaf reflectance spectra: algorithm development for remote sensing of chlorophyll. *Plant Physiol.*, 148:494–500, 1996.

M Govender, P J Dye, I M Weiersbye, E T F Witkowski, and F Ahmed. Review of commonly used remote sensing and ground-based technologies to measure plant water stress. *WaterSA*, 35(5):741–752, 2009.

P. Guo, M. Baum, S. Grando, S. Ceccarelli, G. Bai, R. Li, M. von Korff, R.K. Varshney, A. Graner, and J. Valkoun. Differentially expressed genes between drought-tolerant and drought-sensitive barley genotypes in response to drought stress during the reproductive stage. *Journal of Experimental Botanic*, 60:3531–3544, 2010.

C.A. Hidalgo, N. Blumm, A.L. Barabasi, and N.A. Christakis. A dynamic network approach for the study of human phenotypes. *PLoS Comput Biol*, 5 (4), 04 2009.

M. Hoffman, D.M. Blei, and F. Bach. Online learning for latent dirichlet allocation. In *Proceedings of Neural Information Processing Systems (NIPS-10)*, 2010.

K. Kersting, M. Wahabzada, C. Roemer, C. Thurau, A. Ballvora, U. Rascher, J. Leon, C. Bauckhage, and L. Pluemer. Simplex distributions for embedding data matrices over time. In I. Davidson and C. Domeniconi, editors, *Proceedings of the 12th SIAM International Conference on Data Mining (SDM)*, Anaheim, CA, USA, April 26–28 2012.

C. Lebreton, V. Lazic-Jancic, A. Steed, S. Pekic, and S.A. Quarrie. Identification of qtl for drought responses in maize and their use in testing causal relationships between traits. *Journal of Experimental Botanic*, 46:853–865, 1995.

M.W. Mahoney and P. Drineas. Cur matrix decompositions for improved data analysis. *Proceedings of the National Academy of Sciences of the United States of America (PNAS)*, 106(3):697–703, 2009.

J.K. McKay, J.H. Richards, S. Sen, T. Mitchell-Olds, S. Boles, E.A. Stahl, T. Wayne, and T.E. Juenger. Genetics of drought adaptation in arabidopsis thaliana ii. qtl analysis of a new mapping population , kas-1 x tsu-1. *Evolution*, 62:3014–3026, 2008.

T. Minka. Estimating a Dirichlet distribution. In *A note publically available from the author's homepage.* 2000.

D. Newman, E. Bonilla, and W. Buntine. Improving topic coherence with regularized topic models. In *Proceedings of NIPS*, 2011.

J. Paisley. Two useful bounds for variational inference. Technical report, Department of Computer Science, Princeton University, Princeton, NJ, 2010.

J.B. Passioura. Environmental biology and crop improvement. *Functional Plant Biology*, 29:537–554, 2002.

E. Pinnisi. The blue revolution, drop by drop, gene by gene. *Science*, 320:171–173, 2008.

M.A. Rabbani, K. Maruyama H. Abe, M.A. Khan, K. Katsura, Y. Ito, K. Yoshiwara, M. Seki, K. Shinozaki, and K. Yamaguchi-Shinozaki. Monitoring expression profiles of rice genes under cold, drought, and high-salinity stresses and abscisic acid application using cdna microarray and rna gel-blot analyses. *Plant Physiology*, 133:1755–1767, 2010.

U. Rascher and R. Pieruschka. Spatio-temporal variations of photosynthesis: The potential of optical remote sensing to better understand and scale light use efficiency and stresses of plant ecosystems. *Precision Agriculture*, 9:355–366, 2008.

U. Rascher, C.L. Nichol, C. Small, and L. Hendricks. Monitoring spatio-temporal dynamics of photosynthesis with a portable hyperspectral imaging system. *Photogrammetric Engineering and Remote Sensing*, 73:45–56, 2007.

R.A. Richards, G.J. Rebetzke, M. Watt, A.G. Condon, W. Spielmeyer, and R. Dolferus. Breeding for improved water productivity in temperate cereals: phenotyping, quantitative trait loci, markers and the selection environment. *Functional Plant Biology*, 37(2):85–97, 2010.

C. Römer, K. Bürling, T. Rumpf, M. Hunsche, G. Noga, and L. Plümer. Robust fitting of fluorescence sprectra for presymptomatic wheat leaf rust detection with Support Vector Machines. *Computers and Electronics in Agriculture*, 79(1):180–188, 2010.

T. Rumpf, A.-K. Mahlein, U. Steiner, E.-C. Oerke, and L. Plümer. Early Detection and Classification of Plant Diseases with Support Vector Machines Based on Hyperspectral Reflectance. *Computers and Electronics in Agriculture*, 74(1):91–99, 2010.

S. Seager, E.L. Turner, J. Schafer, and E.B. Ford. Vegetations red edge: A possible spectroscopic biosignature of extraterrestrial plants. *Astrobiology*, 5(3): 372–390, 2005.

J. Sun, Y. Xie, H. Zhang, and C. Faloutsos. Less is more: Sparse graph mining with compact matrix decomposition. *Statistical Analysis and Data Mining*, 1(1):6–22, 2008.

C. Thurau, K. Kersting, M. Wahabzada, and C. Bauckhage. Descriptive matrix factorization for sustainability: Adopting the principle of opposites. *Journal of Data Mining and Knowledge Discovery*, 24(2):325—354, 2012.

M. Wahabzada and K. Kersting. Larger residuals, less work: active document scheduling for latent dirichlet allocation. In *Proceedings of ECML PKDD*, 2011.